\documentclass[11pt,a4paper]{article}
\usepackage[hyperref]{acl2020}
\usepackage{times}
\usepackage{latexsym}

\usepackage{adjustbox}
\usepackage{amsmath}
\usepackage{graphicx}
\usepackage{algorithm}
\usepackage{algorithmic}
\usepackage{multirow}
\usepackage{caption}
\usepackage{subcaption}
\usepackage{mwe}
\usepackage{soul}
\usepackage{url}
\usepackage{amsfonts}
\usepackage{microtype}

\aclfinalcopy % Uncomment this line for the final submission
\title{Two-Headed Monster And Crossed Co-Attention Networks}

\author{Yaoyiran Li\Thanks{The work was done when Yaoyiran was working at Living Analytics Research Centre, Singapore Management University who is now a PhD student at University of Cambridge.}\ \ , \ \ Jing Jiang\\
Living Analytics Research Centre\\
Singapore Management University\\
  \texttt{yl711@cam.ac.uk, jingjiang@smu.edu.sg}} %\\ %And
\date{}

\begin{document}
\maketitle
\begin{abstract}
This paper presents some preliminary investigations of a new co-attention mechanism in neural transduction models. We propose a paradigm, termed Two-Headed Monster (THM), which consists of two symmetric encoder modules and one decoder module connected with co-attention. As a specific and concrete implementation of THM, Crossed Co-Attention Networks (CCNs) are designed based on the Transformer model. We demonstrate CCNs on WMT 2014 EN-DE and WMT 2016 EN-FI translation tasks and our model outperforms the strong Transformer baseline by 0.51 (big) and 0.74 (base) BLEU points on EN-DE and by 0.17 (big) and 0.47 (base) BLEU points on EN-FI.
\end{abstract}

\section{Introduction}
Attention has emerged as a prominent neural module extensively adopted in a wide range of deep learning research problems ~\cite{das2017human,hermann2015teaching,rocktaschel2015reasoning,santos2016attentive,xu2016ask,yang2016stacked,yin2016abcnn,zhu2016visual7w,xu2015show,chorowski2015attention} such as VQA, reading comprehension, textual entailment, image captioning, speech recognition and so forth. It's remarkable success is also embodied in machine translation tasks ~\cite{bahdanau2014neural, 46201}. 

\begin{figure}[tbh]
\centering
\includegraphics[width=5cm]{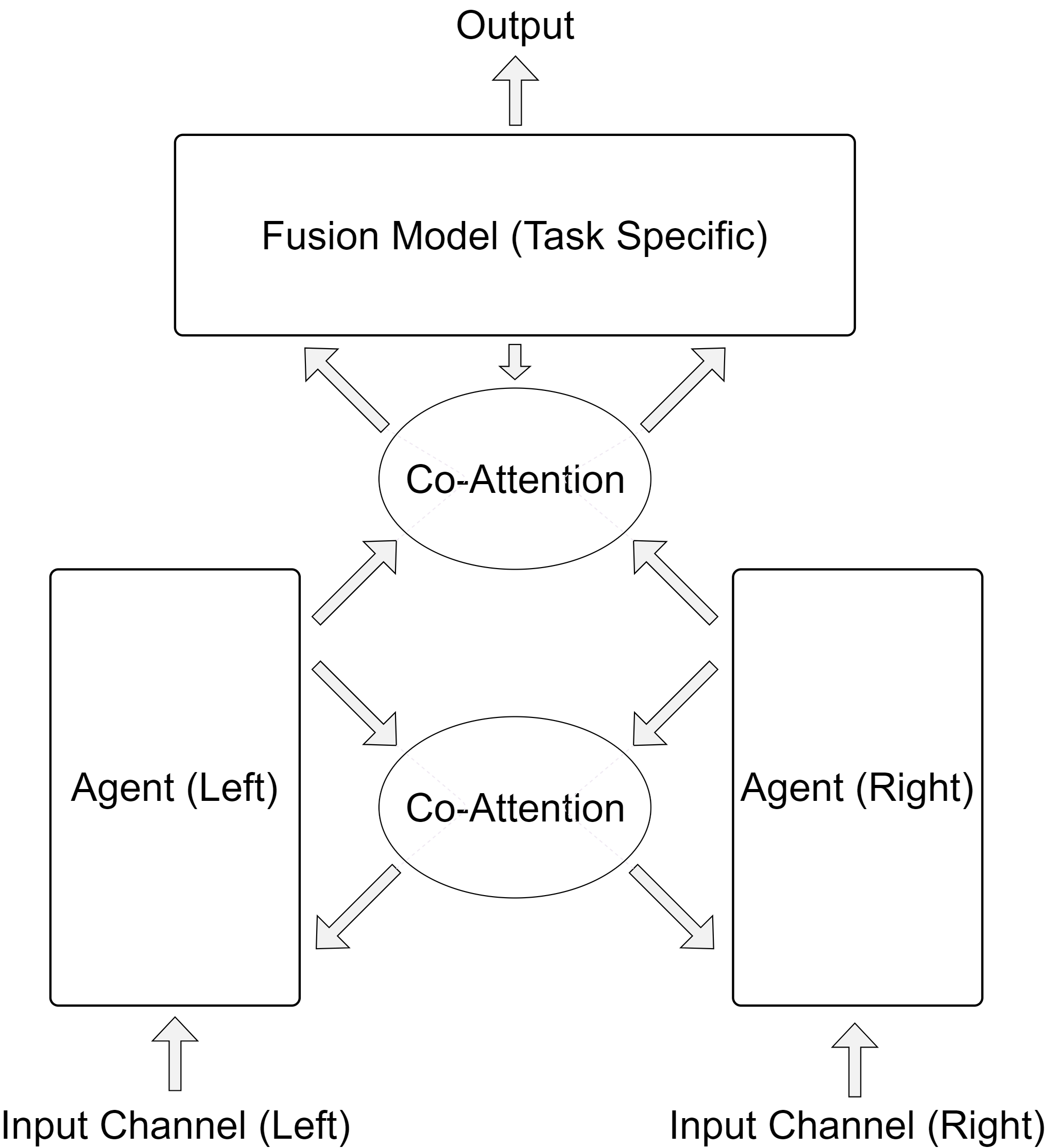}
\caption{Two-Headed Monster.}
\label{fig:thm}
\end{figure}

This work proposes an end-to-end co-attentional neural structure, named Crossed Co-Attention Networks (CCNs) to address machine translation, a typical sequence-to-sequence NLP task. We customize the transformer ~\cite{46201} featured by non-local operations ~\cite{NonLocal2018} with two input branches and tailor the transformer's multi-head attention mechanism to the needs of information exchange between these two parallel branches. A higher-level and more abstract paradigm generalized from CCNs is denoted as "Two-Headed Monster" (THM), representing a broader class of neural structure benefiting from two parallel neural channels that would be intertwined with each other through, for example, co-attention mechanism as illustrated in Fig. \ref{fig:thm}. 
Needless to say, co-attention is widely adopted in multi-modal scenarios ~\cite{NIPS2016_6202,yu2017multi,tay2018multi,xiong2016dynamic,lu2016hierarchical}, the basic idea of which is to make two feature maps from different domains to attend to each other symmetrically and thus output summarized representations for each domain. In this work, we emphasize a parallel and symmetric manifold operating on two input channels and possessing two output channels but do not assume that the two channels of input must be disparate. Our co-attention mechanism is designed in a "Transformer" style, and to the best of our knowledge, our proposed Crossed Co-Attention Network is one of the first (if not the only) implementations of co-attention on transformer model. As a preliminary investigation, we apply our model on the popular machine translation task where two input channels are in one same domain. Our code also leverages half-precision floating point format (FP16) training and synchronous distributed training for inter-GPU communication (we do not discard gradients calculated by "stragglers") which dramatically accelerate our training procedure ~\cite{ott2018scaling,micikevicius2018mixed}. We will release our code after the paper is de-anonymized.
\section{Model Architecture}
We propose an end-to-end neural architecture, based on the transformer, to address a class of sequence to sequence tasks where the model takes input from two channels. We design a Crossed Co-Attention Mechanism to make our model capable of attending to two information flows simultaneously in both the encoding and the decoding stages. Our co-attention mechanism is naively realized by a crossed connection of Value, key and Query gates of a regular multi-head attention module, so we term our model Crossed Co-Attention Networks.
\subsection{Generic Co-Attention}
In this section, we first review non-local operations and bridge them to the dot-product attention that is widely used in self-attention modules and then formulate the co-attention mechanism in a generic way. A non-local operation is defined as a building block in deep neural networks which captures long-range dependencies where every response is computed as a linear combination of all features in the input feature map ~\cite{NonLocal2018}. Suppose the input feature maps are $ V = [v_{1}, v_{2}, ..., v_{n}]^{T}\in \mathbb{R}^{n \times d}$, $ K = [k_{1}, k_{2}, ..., k_{n}]^{T}\in \mathbb{R}^{n \times d}$ and $ Q = [q_{1}, q_{2}, ..., q_{n}]^{T}\in \mathbb{R}^{n \times d}$ and the output feature map $ Y = [y_{1}, y_{2}, ..., y_{n}]^{T}\in \mathbb{R}^{n \times d}$ is of the same size as the input. Then a generic non-local operation is formulated as follows:
\begin{equation}
\it{y}_{i} =\frac{1}{\it{C} (\it{q_{i},K})}\sum_{\forall j} \it{f}(\it{q}_i , \it{k}_{j}) \it{g} (\it{v}_{j})
\label{formula:1}
\end{equation}
We basically follow the definition of no-local operation in ~\cite{NonLocal2018} where $\it{f}:\mathbb{R}^{d} \times \mathbb{R}^{d} \rightarrow \mathbb{R}$ is a pairwise function ("$\times$" is Cartesian product),  $\it{g}:\mathbb{R}^{d} \rightarrow \mathbb{R}^{d}$ is a unary function and $\it{C}:\mathbb{R}^{d} \times \mathbb{R}^{n \times d} \rightarrow \mathbb{R}$ calculates a normalizer, but dispense with the assumption that $\it{V} = \it{K} = \it{Q}$. However, if we assume $f(\it{q_{i}} , \it{k_{j}}) = e^{(q_{i}^{T}W^{Q}) \cdot (k_{j}^{T} W^{K})^{T}}$,  $\it{g(v_{i})=v_{i}^{T}W^{V}}$, the normalizer $\it{c}(\it{q_{i},K})=\sum_{\forall j}f(q_{i},k_{j})$ and $\it{V} = \it{K} = \it{Q}$, then the non-local operation degrades to the multi-head self-attention as is described in ~\cite{46201} (formula \ref{formula:2} describes only one attention head):  
\begin{equation}
\it{Y} = softmax(QW^{Q}(KW^{K})^{T})VW^{V}
\label{formula:2}
\end{equation}
Considering two input channels, denoted as 'left' and 'right', we present the following non-local operation as a definition of co-attention where $\alpha(\cdot), \beta(\cdot) \in \{'left','right'\}$. Note that when $\alpha(\cdot) = 'left', \beta(\cdot) = 'right'$ the co-attention degrades to two self-attention modules. 
\begin{equation}
\begin{aligned}
\it{y}^{left}_{i} &=\frac{1}{\it{C}^{left} (\it{q^{\alpha(Q)}{i},K^{\alpha(K)}})}\cdot \\ \sum_{\forall j}& \it{f}^{left}(\it{q^{\alpha(Q)}}_i , \it{k^{\alpha(K)}}_{j}) \it{g}^{left} (\it{v^{\alpha(V)}}_{j})
\end{aligned}
\label{formula:3}
\end{equation}
\begin{equation}
\begin{aligned}
\it{y}^{right}_{i} &=\frac{1}{\it{C}^{right} (\it{q^{\beta(Q)}{i},K^{\beta(K)}})}\cdot \\ \sum_{\forall j}& \it{f}^{right}(\it{q^{\beta(Q)}}_i , \it{k^{\beta(K)}}_{j}) \it{g}^{right} (\it{v^{\beta(V)}}_{j})
\end{aligned}
\end{equation}
\subsection{Crossed Co-Attention Networks}
Based on the transformer model ~\cite{46201}, we design a novel co-attention mechanism. Our proposed mechanism consists of two symmetrical branches that work in parallel to assimilate information from two input channels respectively. Different from previously known co-attention mechanisms such as ~\cite{xiong,NIPS2016_6202}, our co-attention is built through connecting two multiplicative attention modules ~\cite{46201} each containing three gates, i.e., Value, Key and Query. The information flows from two input channels then interact with and benefit from each other via crossed connections. Suppose the input fed into the left branch is $X_{Left}$, and the right branch $X_{right}$. In our encoder, the left branch takes input from $X_{Left}$ as Value (V) and Key (K) and takes the input $X_{right}$ as Query (Q). The right branch, however, takes the input $X_{Left}$ as Query (Q) and $X_{right}$ as Value (V) and Key (K). This design is, in a sense, meant for the two branches to relatively keep the information in their own domains. A special case is, if $\it{g(v_{i})=v_{i}}$, then the response $y_{i}$ will be in the row space of $V$. Because when an attention takes input $V$ from its own branch, the output responses will by and large carry the information of the branch. For machine translation, the two encoder branches take in one same input sequence, but in order to reduce the redundancy of two parallel branches, we apply dropout and input corruption on input embeddings for two branches respectively. While our model shares BPE embeddings ~\cite{sennrich2015neural} globally, for input matrices encoder branches, we randomly select and swap two sub-word tokens at a probability of $0.5$. 
\begin{figure}[tbh]
\centering
\includegraphics[width=7cm]{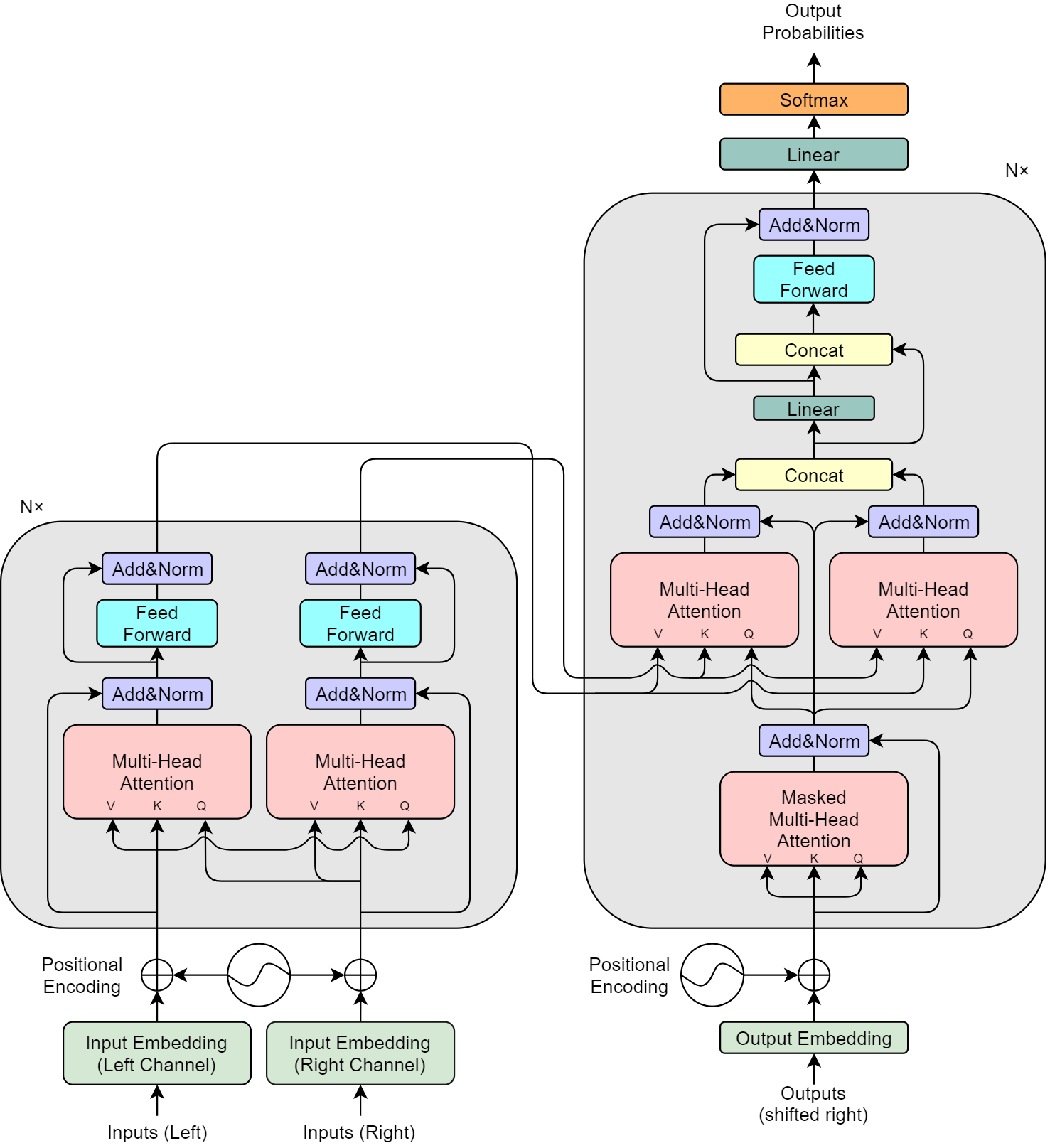}
\caption{Crossed Co-Attention Networks.}
\label{fig:ccn}
\end{figure}
In the encoder-decoder attention layers, the multi-head attention on two decoder branches uses the output from two encoder branches as Value and Key alternatively while absorbing the self-attended output embedding from below as Query. The output of the two branches in decoder is processed through concatenation, linear transformation and then fed into a feed-forward network. In addition to our co-attention mechanism, we keeps one self-attention layer in the decoder for reading in shifted output embedding. We adopt the same input masking and sinusoidal position encoding as the Transformer which will not be expanded here.
\begin{table*}[t]
\begin{center}
\resizebox{\textwidth}{!}{%
\begin{tabular}{llllll}
\hline \multicolumn{1}{c}{\bf Model}  &\multicolumn{1}{c}{\bf Dataset}  &\multicolumn{1}{c}{\bf Epoch Time (s)} &\multicolumn{1}{c}{\bf BLEU} &\multicolumn{1}{c}{\bf Number of Parameters} &\multicolumn{1}{c}{\bf Batch Size}
\\ \hline
\multicolumn{1}{c}{ Transformer-Base}  &\multicolumn{1}{c}{ WMT2014 EN-DE}  &\multicolumn{1}{c}{\bf 684.52} &\multicolumn{1}{c}{ 27.21}  &\multicolumn{1}{c}{ 61,364,224} &\multicolumn{1}{c}{ 6,528}\\
\multicolumn{1}{c}{ THM / CCN-Base}  &\multicolumn{1}{c}{ WMT2014 EN-DE}  &\multicolumn{1}{c}{ 1090.65} &\multicolumn{1}{c}{\bf 27.95} &\multicolumn{1}{c}{ 114,928,640} &\multicolumn{1}{c}{ 6,528}\\
\multicolumn{1}{c}{ Transformer-Base}  &\multicolumn{1}{c}{ WMT2016 EN-FI}  &\multicolumn{1}{c}{\bf 232.97} &\multicolumn{1}{c}{ 16.12}  &\multicolumn{1}{c}{ 55,883,776} &\multicolumn{1}{c}{ 6,528}\\
\multicolumn{1}{c}{ THM / CCN-Base}  &\multicolumn{1}{c}{ WMT2016 EN-FI}  &\multicolumn{1}{c}{ 410.79} &\multicolumn{1}{c}{\bf 16.59}  &\multicolumn{1}{c}{ 109,448,192} &\multicolumn{1}{c}{ 6,528}\\
\multicolumn{1}{c}{ Transformer-Big}  &\multicolumn{1}{c}{ WMT2014 EN-DE}  &\multicolumn{1}{c}{\bf 1982.63} &\multicolumn{1}{c}{ 28.13} &\multicolumn{1}{c}{210,808,832} &\multicolumn{1}{c}{ 2,176}\\
\multicolumn{1}{c}{ THM / CCN-Big}  &\multicolumn{1}{c}{ WMT2014 EN-DE}  &\multicolumn{1}{c}{ 3611.53} &\multicolumn{1}{c}{\bf 28.64} &\multicolumn{1}{c}{424,892,416} &\multicolumn{1}{c}{ 2,176}\\
\multicolumn{1}{c}{ Transformer-Big}  &\multicolumn{1}{c}{ WMT2016 EN-FI}  &\multicolumn{1}{c}{\bf 726.51} &\multicolumn{1}{c}{ 16.21} &\multicolumn{1}{c}{ 199,847,936} &\multicolumn{1}{c}{ 2,176}\\
\multicolumn{1}{c}{ THM / CCN-Big}  &\multicolumn{1}{c}{ WMT2016 EN-FI}  &\multicolumn{1}{c}{ 1387.22} &\multicolumn{1}{c}{\bf 16.38} &\multicolumn{1}{c}{ 413,931,520} &\multicolumn{1}{c}{ 2,176}\\
\hline
%\multicolumn{1}{c}{\bf PART}  &\multicolumn{1}{c}{\bf DESCRIPTION}
%\\ \hline \\
%Dendrite         &Input terminal \\
%Axon             &Output terminal \\
%Soma             &Cell body (contains cell nucleus) \\
\end{tabular}
}
\caption{\label{font-table} Comparisons Between Our Proposed Method and Transformer Baseline on WMT 2014 EN-DE and WMT 2016 EN-FI}
\end{center}
\end{table*}
\section{Experiments}
\subsection{Setup}
We demonstrate our model on WMT 2014 EN-DE and WMT 2016 EN-FI machine translation tasks. For convenience, in this section, we do not differentiate between the notion of THM and CCN which is an implementation of THM. The raw input data is pre-processed with length filtering as previous work ~\cite{ott2018scaling}. Our final dataset consists of $4,575,637$ training examples, $3,000$ valid examples and $3,003$ test examples for EN-DE, and $2,073,194$ training examples, $1,500$ valid examples and $3,000$ test examples for EN-FI. Considering the scale of the training sets, we adopt shared BPE dictionaries of size $33,712$ for EN-DE and $23,008$ for EN-FI. Our CCNs are established with $6$ encoder and decoder blocks and a hidden state of size $512$ for base models and with also $6$ such blocks but a hidden state of $1,024$ neurons for big models. That exactly corresponds to the settings of Transformer paper. We train our models on a NVIDIA DGX-1 GPU server with $4$ TESLA V100-16GB GPUs. In order to make full use of the computational resources, FP16 computation is adopted and we use a batch size of $6,528$ tokens/GPU for base models and $2,176$ for big models (both Transformer and THM). We adopt the Sequence-to-Sequence Toolkit FairSeq ~\cite{ott2019fairseq} released by the Facebook AI Research for our Transformer baseline \footnote{https://github.com/pytorch/fairseq}, upon which our THM code is built as well.  We train all base models for around one day and big models for around two days. For model selection, we strictly choose the model that achieves the highest BLEU on Dev set.
\subsection{Experimental Results}
\paragraph{Main Results:}
Our experiments demonstrate the efficiency of our proposed crossed co-attention mechanism which significantly improves the BLEU scores of machine translation as illustrated in Table ~\ref{font-table}. Besides, the co-attention mechanism has, by and large, reduced training, valid and test loss from the first training epoch compared with the transformer baselines as shown in Appendices ~\ref{sec:appendix1}. However, since the number of parameters doubles, the epoch time also increases by roughly $60\% \sim 80\%$. 
\paragraph{Capability of Model Selection:}
In addition to the BLEU, loss and time efficiency, we also find that the THM/CCN models demonstrate better capability of selecting good models with Dev set from all models derived in all training epochs. As is shown is Table ~\ref{font-table2}, for THM/CCN, the models that achieved hightest BLEU on Dev set are also high-ranking on the Test set. In $75\%$ cases, THM will select TOP $3$ models and in all cases, it will select TOP $10$ models whereas Transformer can only select TOP $10$ models in $50\%$ cases.   
\paragraph{Performance across Languages:}
We test our proposed method on two language pairs, EN-DE and EN-FI and the improved BLEU scores and the capability of model selection on both base and big models demonstrate the universality of our proposed method.
\begin{table}
\centering
\begin{tabular}{lrl}
\hline & \multicolumn{1}{c}{\bf THM / CCN} & \multicolumn{1}{c}{\bf Transformer} \\ \hline
\multicolumn{1}{c}{TOP 1} & \multicolumn{1}{c}{\bf 25\%} & \multicolumn{1}{c}{0} \\
\multicolumn{1}{c}{TOP 3} & \multicolumn{1}{c}{\bf 75\%}  & \multicolumn{1}{c}{0} \\
\multicolumn{1}{c}{TOP 5} & \multicolumn{1}{c}{\bf 100\%}  & \multicolumn{1}{c}{0}\\
\multicolumn{1}{c}{TOP 10} & \multicolumn{1}{c}{\bf 100\%}  & \multicolumn{1}{c}{50\%} \\
\hline
\end{tabular}
\caption{\label{font-table2} This Table Evaluates If The Models Selected by The Dev Set Are Also Better Than Others on Test Set. Here We Provide The Percentage of Selected Models That Rank TOP 1, TOP 3, TOP 5 or TOP 10 Among All Models Derived from All Training Epochs.}
\end{table}
\section{Related Work}
\paragraph{Attention:}
Multi-head self-attention has demonstrated its capacity in neural transduction models ~\cite{46201}, language model pre-training ~\cite{devlin2018bert,radford2018improving} and speech synthesis ~\cite{yang2019enhancing}. While the novel attention mechanism, eschewing recurrence, is famous for modeling global dependencies and considered faster than recurrent layers ~\cite{46201}, recent work points out that it may tend to overlook neighboring information ~\cite{yang2019context,xu2019leveraging}. It is found that applying an adaptive attention span could be conducive to  character level language modeling tasks ~\cite{sukhbaatar2019adaptive}. Yang et al. propose to model localness for self-attention which would be conducive to capturing local information by learning a Gaussian bias predicting the region of local attention ~\cite{yang2018modeling}. Other work indicates that adding convolution layers would ameliorate the aforementioned issue ~\cite{yang2018glomo,yang2019convolutional}. Multi-head attention can also be used in multi-modal scenarios when V, K and Q gates take in data from different domains. ~\cite{helcl2018cuni} adds an attention layer on top of the encoder-decoder layer with K and V being CNN-extracted image features.
\paragraph{Machine Translation:}
Some recent advances in machine translation aim to find more efficient models based on the Transformer: Hao et al. add an additional recurrence encoder to model recurrence for Transformer~\cite{hao2019modeling}; So et al. demonstrate the power of neural architecture search and find that the found evolved transformer architecture outperforms human-designed ones ~\cite{so2019evolved}; Wu et al. propose dynamic convolutions that would be more efficient and simpler compared with self-attention ~\cite{wu2019pay}. Other work shows that training on $128$ GPUs can significantly boost the experimental results and shorten the training time ~\cite{ott2018scaling}. A novel research direction is semi- or un-supervised machine translation aimed at addressing low-resource languages where parallel data is usually unavailable  ~\cite{cheng2019semi,artetxe2017unsupervised,lample2017unsupervised}.
\section{Conclusion}
We propose a novel co-attention mechanism consisting of two parallel attention modules connected with each other in a crossed manner. First we formulate the co-attention in a general sense as a non-local operation and then show a specific type of co-attention, known as crossed co-attention can improve the machine translation tasks by $ 0.17 \sim 0.74 $ BLEU points and enhance the capability of model selection. However, the time efficiency is reduced since the number of parameters increases. % when we feed the same input sequence to two input branches simultaneously.
%\section*{Acknowledgments}

\bibliography{anthology,acl2020}
\bibliographystyle{acl_natbib}

\appendix

\section{Appendices}
\label{sec:appendix}

\subsection{ Comparisons of Loss between CCN models and Transformer baselines.}
\label{sec:appendix1}
\begin{figure*}[tbh]
\hspace*{-1.638cm}   
\centering
\includegraphics[width=18.66cm]{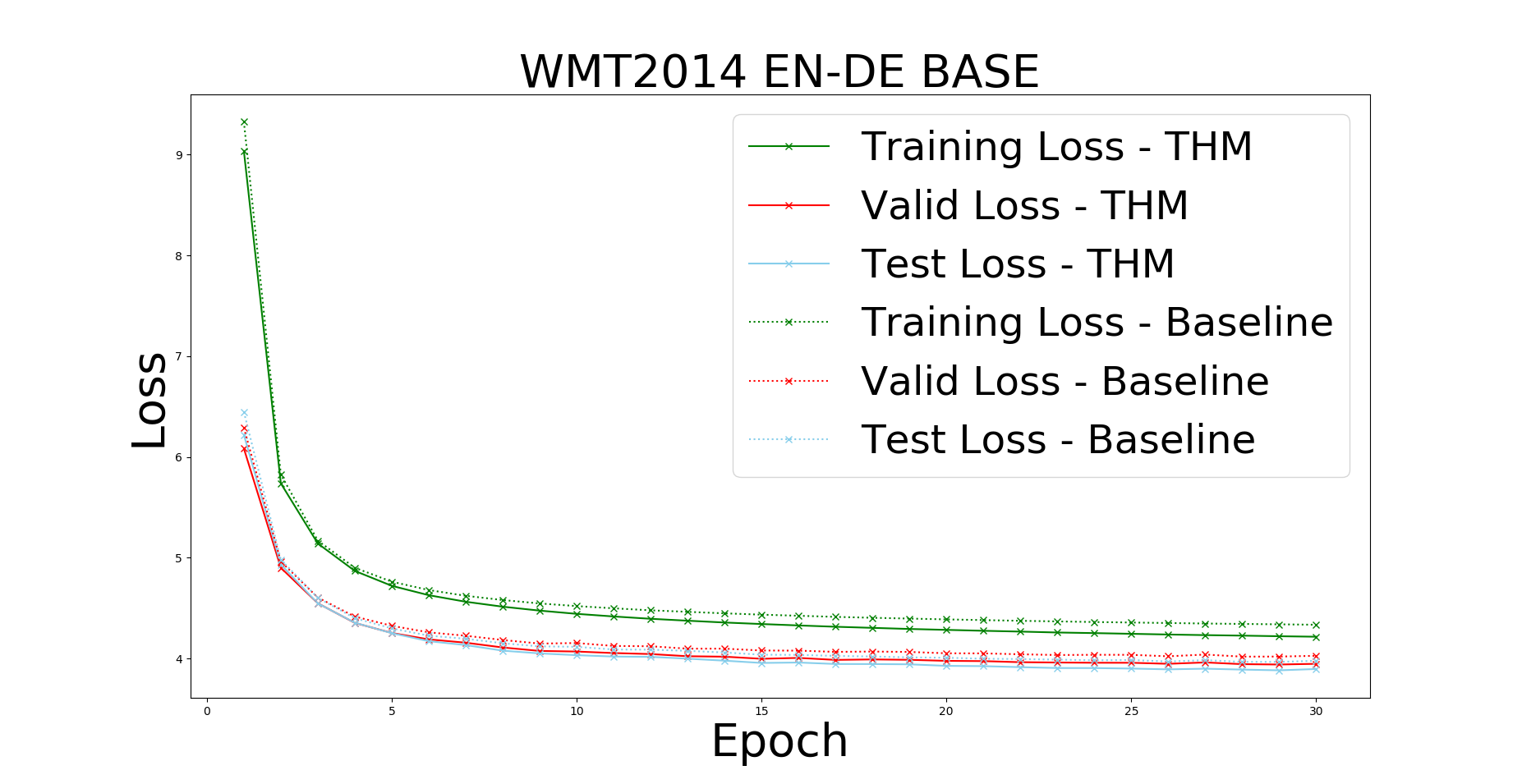}
\caption{Loss vs Epoch for THM-base and Transformer-base on EN-DE}
\label{fig:mean and std of net14}
\end{figure*}

\begin{figure*}[tbh]
\hspace*{-1.638cm}   
\centering
\includegraphics[width=18.66cm]{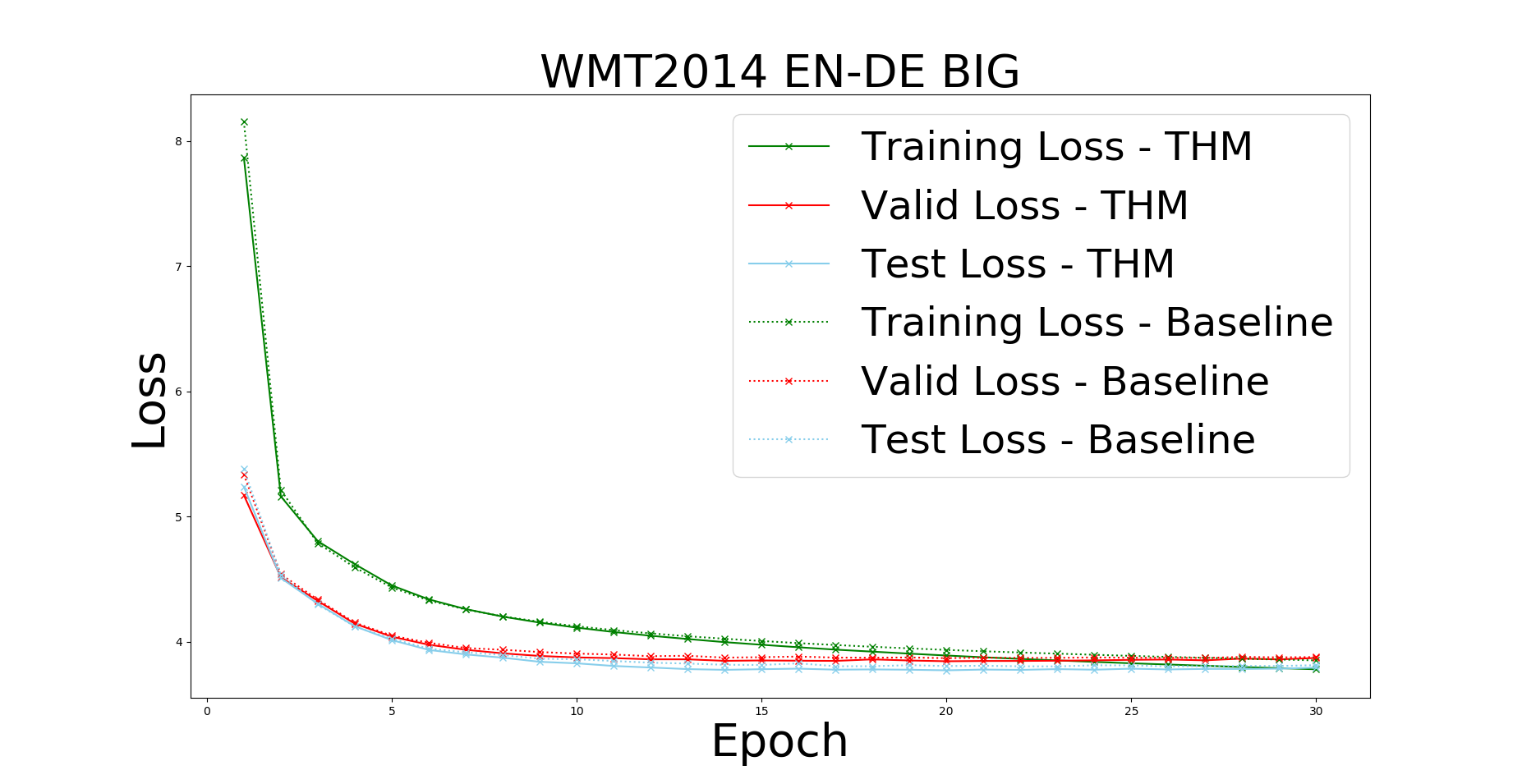}
\caption{Loss vs Epoch for THM-big and Transformer-big on EN-DE}
\label{fig:mean and std of net24}
\end{figure*}

\begin{figure*}[tbh]
\hspace*{-1.638cm}   
\centering
\includegraphics[width=18.66cm]{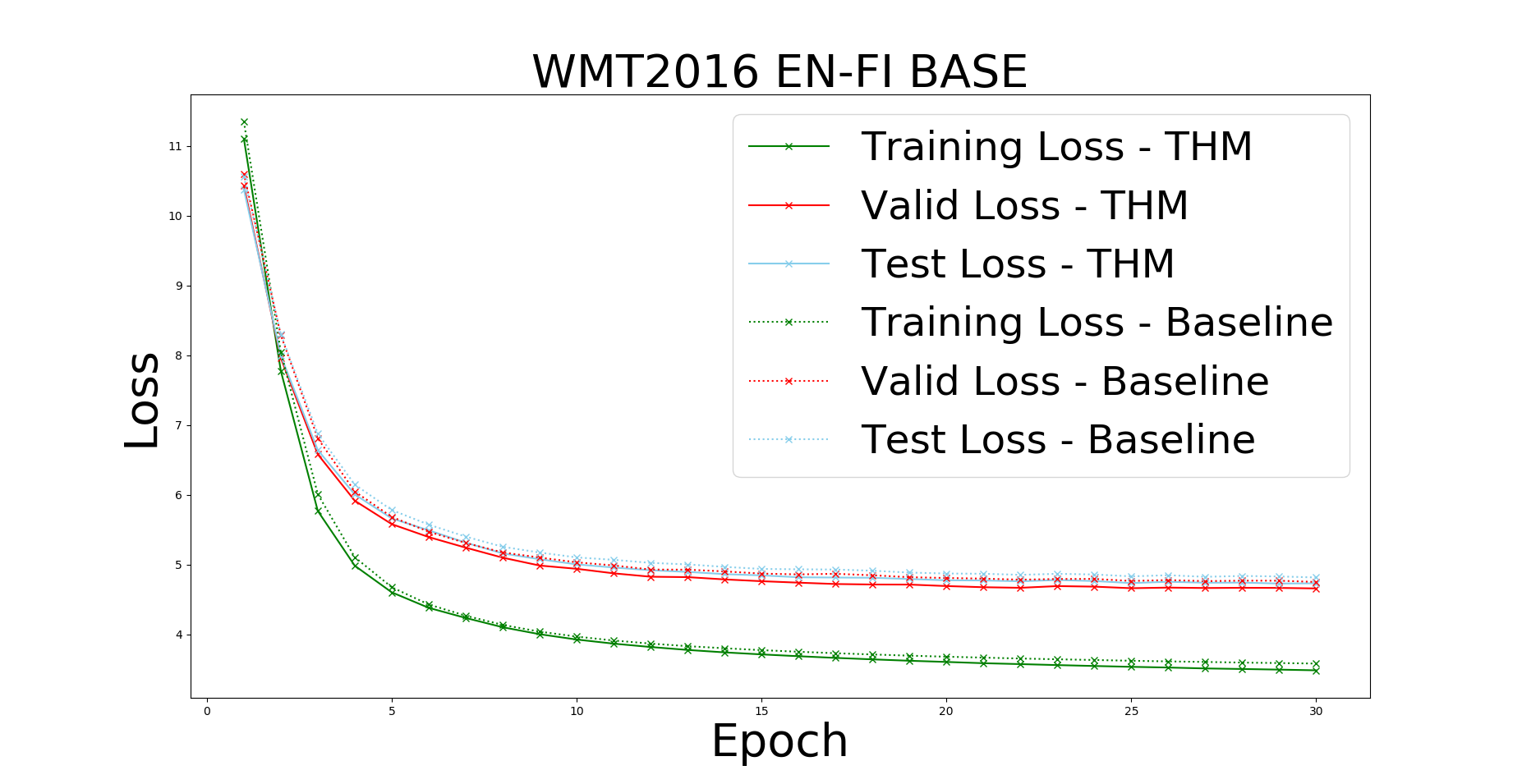}
\caption{Loss vs Epoch for THM-base and Transformer-base on EN-FI}
\label{fig:mean and std of net34}
\end{figure*}

\begin{figure*}[tbh]
\hspace*{-1.638cm}   
\centering
\includegraphics[width=18.66cm]{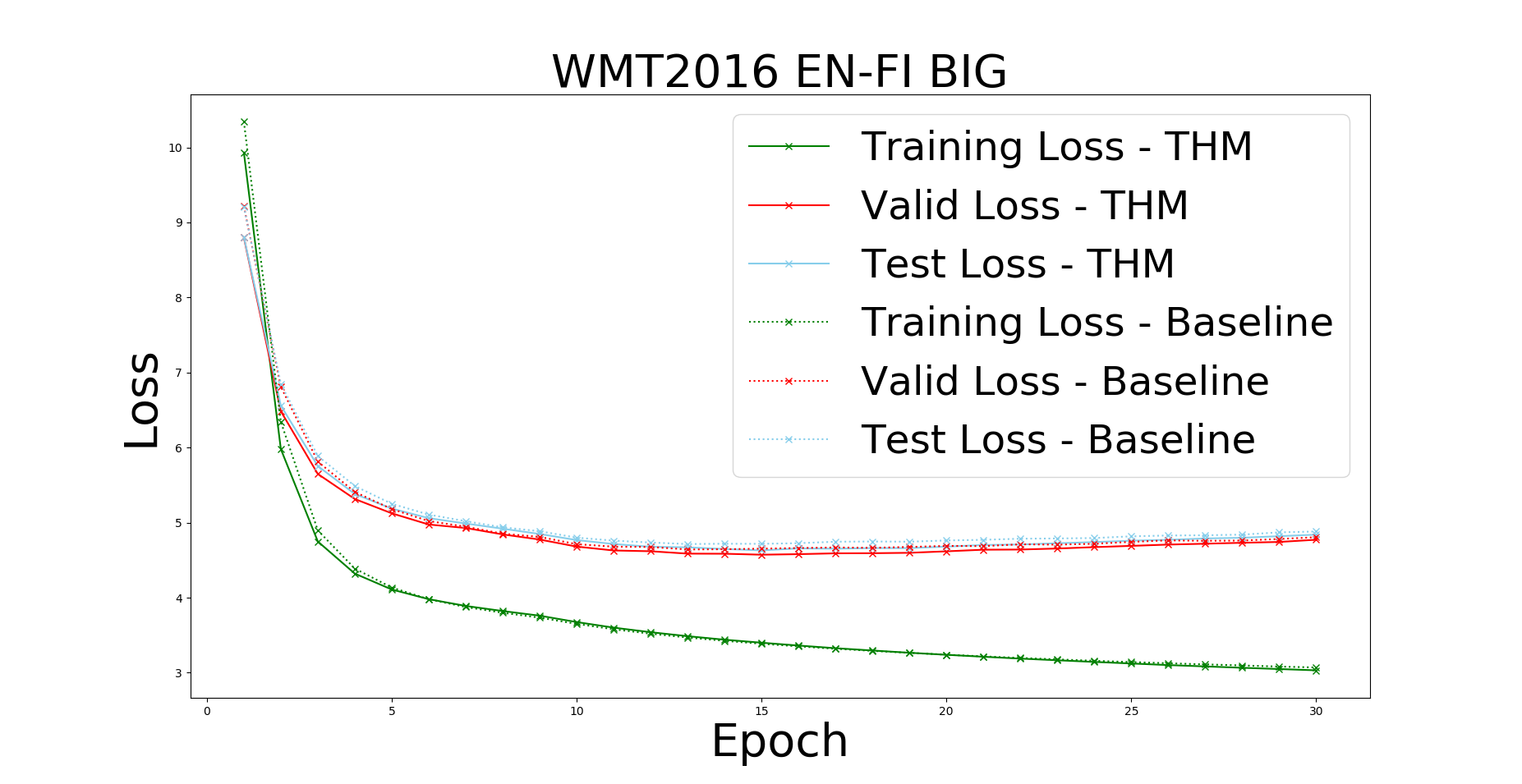}
\caption{Loss vs Epoch for THM-big and Transformer-big on EN-FI}
\label{fig:mean and std of net44}
\end{figure*}

\end{document}